\documentclass[11pt,a4paper]{article}
\usepackage[hyperref]{acl2020}
\usepackage{amsmath}
\usepackage{times}
\usepackage{subfigure}
\usepackage{latexsym}
\usepackage{booktabs}
\usepackage{enumitem}

\usepackage{microtype}

\aclfinalcopy 
\def\aclpaperid{2536} 

\title{On the Spontaneous Emergence of Discrete and Compositional Signals}

\author{%
	Nur Geffen Lan\textsuperscript{1} \quad Emmanuel Chemla\textsuperscript{2} \quad Shane Steinert-Threlkeld\textsuperscript{3}
	\\
	\textsuperscript{1} Computational Linguistics Lab, Tel Aviv University
	\\
	\textsuperscript{2} EHESS, PSL University, CNRS, Ecole Normale Sup\'erieure
	\\
	\textsuperscript{3} Department of Linguistics, University of Washington
	\\
	\texttt{nurlan@mail.tau.ac.il}\quad\texttt{chemla@ens.fr}\quad\texttt{shanest@uw.edu}
}

\date{}

\usepackage{multicol}  
\usepackage{amsmath}
\usepackage{graphicx}

\DeclareMathOperator*{\argmax}{arg\,max}
\DeclareMathOperator*{\argmin}{arg\,min}

\begin{document}

\maketitle

\begin{abstract}
	We propose a general framework to study language emergence through signaling games with neural agents. Using a continuous latent space, we are able to (i)~train using backpropagation, (ii)~show that discrete messages nonetheless naturally emerge.  We explore whether categorical perception effects follow and show that the messages are not compositional. 
\end{abstract}

\section{Introduction}

In a signaling game, artificial agents learn to communicate to achieve a common goal: a sender sees some piece of information and produces a message, which is then sent to a receiver that must take some action \citep{Lewis1969, Skyrms2010}. If the action is coherent with the sender's initial piece of information, the choice of the message and its interpretation is reinforced. 
For instance, in a referential game, sender and receiver see a set of objects, and the sender knows which of these the receiver must pick; the sender then sends a message to the receiver, who must interpret it to pick up the right object \citep{Lazaridou2017, Lazaridou2018, Havrylov2017, Chaabouni2019a}.

This setting has been used to study the factors influencing the emergence of various fundamental properties of natural language, such as \emph{compositionality} \citep{Kirby2015, Franke2016, SteinertThrelkeld2016, Mordatch2018, Lazaridou2018, Choi2018}.
In this paper, we add focus on two other so-called `design features' of natural language \citep{Hockett1960}: 
\emph{discreteness} (i.e.~words form clusters in acoustic space), 
and \emph{displacement} (i.e.~efficient communication can occur about objects and facts beyond the immediate context of the conversation).

From an implementation point of view, we follow the recent literature which has shown that a signaling game is essentially an autoencoder setting, with the encoder playing the role of the sender, and the decoder the role of the receiver (see Fig.~\ref{fig:model}). In this literature, however, the discreteness of the communication protocol is assumed, since the networks then traditionally use a (normally sequential and) discrete latent space \citep{Havrylov2017, Chaabouni2019a, Kharitonov2019}.

Our main contribution is a generalization of the current implementation of signaling games as autoencoders.
Our implementation covers a broader variety of signaling games, and it crucially incorporates the possibility of displacement and makes no \emph{a priori} assumption of discreteness.
Our main result is that under appropriate conditions, discreteness emerges spontaneously: if the latent space is thought about as a continuous acoustic space, then trained messages form coherent clusters, just like regular words do.  We also show that the messages are not compositional.

In addition to contributing to our understanding of the emergence of communication protocols with features like natural language, our results have technical significance: by using a continuous communication protocol, with discreteness spontaneously emerging, we can train end-to-end using standard backpropagation, instead of reinforcement learning algorithms like REINFORCE and its refinements \citep{Williams1992, Schulman2015, Mnih2016}, which are difficult to use in practice.

\section{Related Work}

A related line of work attempts to avoid the difficulties of reinforcement learning---used when there are stochastic nodes in a computation graph---by reparameterization and/or non-stochastic estimators \citep{Bengio2013, Schulman2015}.  In the emergent communication case, where the stochastic nodes are discrete (e.g.\ sampling a message from a sender distribution), the Gumbel-Softmax estimator has become increasingly popular \citep{Jang2017, Maddison2017}.  

That work enables standard backpropagation to be used for training by optimizing approximations to the true reinforcement learning signal.  By contrast, we do not approximate the discrete RL learning signal, but rather ask under what conditions discreteness will emerge. 

Several earlier papers explore similar topics in the emergence of discrete symbols.  \citet{Nowak1999a} show that the division of the acoustic space is an emergent property of language use under noise.
It assumes that speakers have a fixed language and asks which such ones are stable.  In our setting, the language itself is changing as the result of reinforcement from communication and transmission itself is not noisy.  

\citet{DeBoer2000} simulates the emergence of vowel systems in artificial agents modeled after phonetic production and perception in humans, resulting in a self-discretizing acoustic space and a vowel system that resembles human ones.
This makes the agents much closer to what we know about humans, but also limits its scope.  Results about emergent communication can tell us both about the emergence of human language, but also about communication protocols in general, that may be used by very different agents, e.g.\ autonomous ones, or animals \cite{Steinert-Threlkeld2020a}.

\section{Function Games}

We here introduce a general communication game setting, which we call Function Games.  Our games contain three basic components: (i) a set of contexts $C$, (ii) a set of actions $A$, (iii) a family of functions $F$, from contexts to actions.  One play of a Function Game game runs as follows:
\begin{enumerate}[noitemsep]
	\item Nature chooses $f \in F$ and a context $c \in C$.
	\item Sender sees the context $c$ and $f$.
	\item Sender sends a message $m$ to Receiver.
	\item Receiver sees \emph{a possibly different} context $c'$ and the message $m$ and chooses an action $a'$.
	\item Both are `rewarded' iff $a' = f(c')$.
\end{enumerate}
Abstractly, the function $f$ represents some piece of knowledge available primarily for Sender, and which determines what action is appropriate in any given context. Two concrete interpretations will help illustrate the variety of communication protocols and goals that this framework encompasses.

\noindent \textbf{Generalized referential games.}  A reference game is one in which Sender tries to get Receiver to pick the correct object out of a given set \citep{Skyrms2010, Lazaridou2017, Lazaridou2018, Havrylov2017, Chaabouni2019a}.  Here, contexts are sets of objects (i.e.\ an $m \times n$ matrix, with $m$ objects represented by $n$ features).  Normally (though we will drop this assumption later), $c' = \texttt{shuffled}(c)$: Sender and Receiver see the same objects, but in a different arrangement. Actions are the objects, and the functions $f \in F$ are \emph{choice functions}: $f(c) \in c$ for every context $c$.

\noindent \textbf{Belief update games.} We will mostly focus on the previous interpretation, but illustrate the generality of the setting with another interpretation here. Contexts can represent the (possibly different) belief states of the agents. `Actions' can represent updated belief states ($A = C$), the different functions in $F$ then representing how to update an agent's beliefs in the light of learning a particular piece of information (passed directly to Sender, and only through the message to Receiver). 

\section{Experiment}

Because we are interested in the simultaneous emergence both of discrete and of compositional signals, we use a Function Game called the Extremity Game designed to incentivize and test rich compositionality \citep{Steinert-Threlkeld2019, Steinert-Threlkeld2020}. In this game, one may think of the $n$ dimensions of the objects as gradable properties, e.g. size and darkness, so that a 2D object is determined by a given size and shade of gray. For the functions, we set $F = \left\{ \argmin_i , \argmax_i : 0 \leq i < n \right\}$. An emerging language may contain compositional messages like `\textsc{most} + \textsc{big}', `\textsc{least} + \textsc{dark}'.

\subsection{Model}

\begin{figure}[ht]
	\centering
	\includegraphics[width=\columnwidth, trim=0 150 0 40]{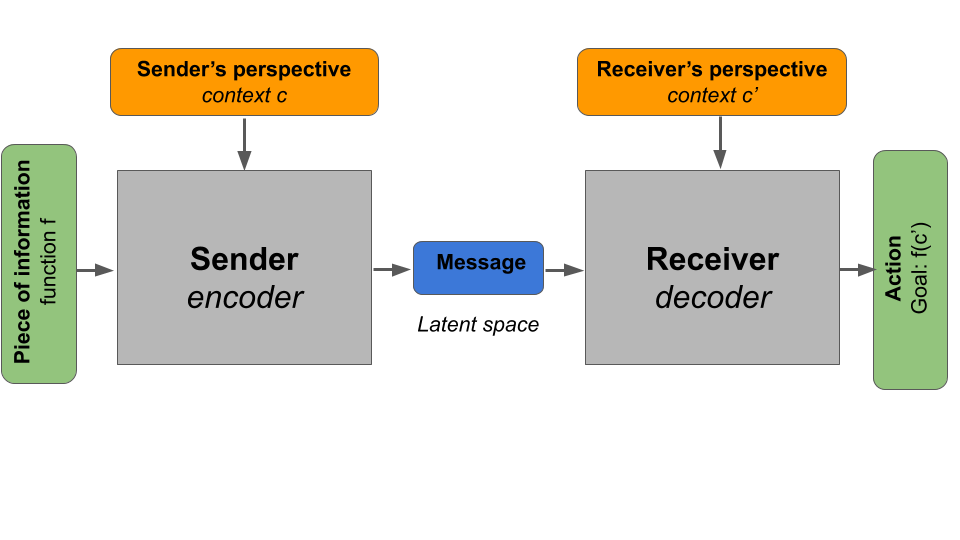}
	\caption{Our model architecture, mixing terminology from the autoencoder and signaling game traditions.}
	\label{fig:model}
\end{figure}

Our model (Figure~\ref{fig:model}) resembles an encoder-decoder architecture, with Sender encoding the context/target pair into a message, and Receiver decoding the message (together with its context $c'$) into an action.  Both the encoder and decoder are multi-layer perceptrons with two hidden layers of 64 ReLU units \citep{Nair2010, Glorot2011}. A smaller, intermediate layer without an activation function bridges the encoder and decoder and represents the transformation of the input information to messages. 

\subsection{Game Parameters}

We manipulate the following parameters:
\setlist{nolistsep}
\begin{itemize}[noitemsep]
	\item Context identity. 
	In the \emph{shared} setting, Receiver sees a shuffled version of Sender's context ($c' = \texttt{shuffled}(c)$). In the \emph{non-shared} setting, Receiver's context $c'$ is entirely distinct from Sender's. This forces displacement and may incentivize compositional messages, since Sender cannot rely on the raw properties of the target object in communication.

	\item Context strictness. 
	In \emph{strict} contexts, there is a one-to-one (and onto) correspondence between $F$ and $A$ (as in the original Extremity Game from \citealp{Steinert-Threlkeld2019, Steinert-Threlkeld2020}). 
In \emph{non-strict} contexts, an object may be the $\argmax$ or $\argmin$ of several dimensions, or of no dimension.

\end{itemize}
In all experiments, the latent space (message) dimension is always 2, and objects have 5 dimensions. Strict contexts therefore contain 10 objects, while non-strict contexts contain 5, 10, or 15 objects.

\subsection{Training Details}

We use the Adam optimizer \citep{Kingma2015} with learning rate 0.001, $\beta_1 = 0.9$, and $\beta_2 = 0.999$. The model is trained for 5,000 steps by feeding the network mini-batches of 64 contexts concatenated with one-hot function selectors. The network's loss is taken as the MSE between the target object $f(c')$ and the object generated by the Receiver. For each setting of the above parameters, we run 20 trials with different random seeds.\footnote{The project's code for extension and reproduction is available at \mbox{\href{https://github.com/0xnurl/signaling-auto-encoder}{https://github.com/0xnurl/signaling-auto-encoder}}.}

\section{Results}

\subsection{Communicative success}
\label{subsection:communicative_success}

We measure the communicative success of the network by calculating the accuracy of recovering the correct object from $c'$. Receiver's prediction is considered correct if its output is closer to $f(c')$ than to all other objects in $c'$. Accuracy of the different settings is reported in Table~\ref{tab:object_prediction_accuracy}. While the network handles displacement well (\textit{non-shared contexts}), the model struggles with non-strict contexts. Note that although accuracy is not $100\%$, it is still well above chance, since e.g. for a context of 10 objects random guessing yields an expected accuracy of $10\%$ (which we observe in our model before training).

\begin{table}[]
\begin{tabular}{lcc}
\toprule
                       & \textbf{Shared} & \textbf{Non-shared} \\ \midrule
\textbf{Strict} & \\
\emph{10 objects} & 
	$63.78\%\pm 1.63$  & $60.22\%\pm 1.56$      \\ 
\textbf{Non-strict} & \\
\emph{5 objects} &
	$49.37\%\pm 1.67$  & $43.55\%\pm 1.69$      \\
\emph{10 objects} &
	$33.06\%\pm 1.47$  & $31.89\%\pm 1.63$      \\
\emph{15 objects} &
	$27.58\%\pm 1.30$  & $27.95\%\pm 1.24$      \\ \bottomrule
\end{tabular}
\caption{Communicative success, as measured by object recovery accuracy.}\label{tab:object_prediction_accuracy}
\end{table}

\subsection{Discrete signals}

\begin{figure}[]
\begin{tabular}{cc}
(a)~Before training &
(b)~After training\\
\includegraphics[scale=.3, trim=15 21 50 0, clip]{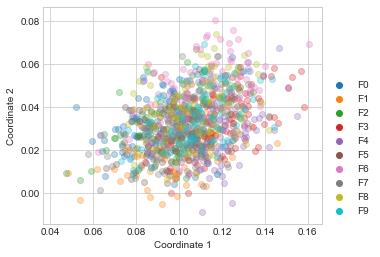} &
\includegraphics[scale=.3, trim=15 21 0 0, clip]{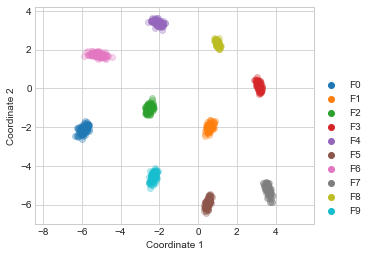}
\end{tabular}
\caption{Sampled messages for contexts of 10 objects of size 5 for (a)~an untrained and (b)~a trained network. Colors represent the $f_i \in F$ input part of the Sender.}
 \label{fig:messages}
\end{figure}

Figure~\ref{fig:messages} depicts message vectors sampled from the latent space layer, before and after training. It is apparent that discrete messages emerge from the imposed learning regime. We measure cluster tendency more quantitatively through two measures, one considering Sender's \emph{production}, and the other Receiver's \emph{perception}. 

First, we sample 100 contexts, and collect the output of the trained encoder for each of these contexts combined with each possible function $f$. We apply an unsupervized clustering algorithm to this set of produced messages (DBSCAN, \citealp{ester_density-based_1996}, with $\epsilon = 0.5$). A label is assigned to each cluster using the ground truth: the label of a cluster is the function $f$ that was most often at the source of a point in this cluster. This allows us to compute F1-scores, which are reported in Table~\ref{tab:f1_scores}.
The model reached near-optimal clusterization measures in 7 out of 8 parameter settings, with the Non-strict, Non-shared context with 5 objects being the exception.

\begin{table}[]
\begin{tabular}{lcc}
\toprule
                       & \textbf{Shared} & \textbf{Non-shared} \\ \midrule
\textbf{Strict} 
	& \phantom{$63.78\%\pm 1.63$}  & \phantom{$60.22\%\pm 1.56$}\\
\emph{10 objects} & 
	$1.00\pm 0.00$  & $0.90\pm 0.09$      \\ 
\textbf{Non-strict} & \\
\emph{5 objects} &
	$0.99\pm 0.02$  & $0.54\pm 0.15$      \\
\emph{10 objects} &
	$1.00\pm 0.00$  & $0.99\pm 0.01$      \\
\emph{15 objects} &
	$1.00\pm 0.00$  & $1.00\pm 0.00$      \\ \bottomrule
\end{tabular}
\caption{Discreteness in production, as measured by F1 scores for automatically clusterized messages.}
\label{tab:f1_scores}
\end{table}

The second approach is akin to studying perception. Given the clusterization of the message space, we sample new messages from each cluster, and test Receiver's perception of these `artificial' messages, which have never been produced by Sender. 
To sample artificial messages, we take the average of 10 messages from a (now labelled) cluster. These artificial messages are fed to Receiver for 100 different contexts. The output object accuracy for these artificial messages is shown in Table~\ref{tab:average_message_accuracy}. The model achieves recovery accuracy similar to when interpreting actual messages.

In sum, we can identify discrete, abstract regions of the latent space corresponding to different functions in the input, just like words form clusters in acoustic space. 

\begin{table}[]
\begin{tabular}{lcc}
\toprule
                       & \textbf{Shared} & \textbf{Non-shared} \\ \midrule
\textbf{Strict} & \\
\emph{10 objects} & 
	$63.39\%\pm 1.45$  & $55.37\%\pm 3.43$      \\ 
\textbf{Non-strict} & \\
\emph{5 objects} &
	$46.94\%\pm 1.70$  & $29.40\%\pm 5.59$      \\
\emph{10 objects} &
	$32.63\%\pm 1.43$  & $31.51\%\pm 1.62$      \\
\emph{15 objects} &
	$28.24\%\pm 1.11$  & $27.94\%\pm 1.20$      \\ \bottomrule
\end{tabular}
\caption{Discreteness in perception, as measured by object recovery accuracy from artificial messages.}
\label{tab:average_message_accuracy}
\end{table}

\subsection{Compositionality}

Our agents are capable of communicating in abstract situations, namely some in which their contexts are different in the first place. This generalizability suggests that the messages may be `compositional'. We here probe for a candidate compositional structure to the latent space, by asking how the messages relate to the structure of the family of functions $F$.

\newcommand{\WE}[1]{\textsc{we}(#1)}
\newcommand{\MESS}[1]{\textsc{m}(#1)}

First, the pioneering \citealp{mikolov_efficient_2013} looks for compositionality at the level of word embeddings (WE) through addition, most classically asking whether  
\WE{queen}=\WE{king}-\WE{man}+\WE{woman}. In the current Game, we can ask whether the messages are related as follows, for any dimensions $i$ and $j$:
\MESS{$c, \argmax_i$}=\MESS{$c, \argmax_j$}-\MESS{$c, \argmin_j$}+\MESS{$c, \argmin_i$}.
For each such pair of object dimensions we calculate the right-hand side of the equation above for 100 contexts, feed it to Receiver, compare Receiver's output to the output that would have been obtained if \MESS{$c, \argmax_i$} (the left-hand side) had been sent in the first place. This leads to important degradation of average communicative success: a drop of at least 24 percentage points across parameter combinations, to around chance level. 
Full results are in the left column of Table~\ref{tab:composition_accuracy}.

\begin{table*}
	\begin{center}
\begin{tabular}{lcccc}
\toprule
                       & \multicolumn{2}{c}{\textbf{Compositionality by Addition}} & \multicolumn{2}{c}{\textbf{Composition Network}} \\ \midrule
                       & \textbf{Shared} & \textbf{Non-shared} & \textbf{Shared} & \textbf{Non-shared} \\ \midrule
\textbf{Strict} & \\ 
\emph{10 objects} &  $7.82\%\pm 2.40$  & $11.94\%\pm 2.13$ &  $13.70\%\pm 6.85$  & $10.18\%\pm 6.15$      \\ 
\textbf{Non-strict} & \\
\emph{5 objects} &
	$16.86\%\pm 3.23$  & $17.14\%\pm 3.54$ &  $15.10\%\pm 2.05$  & $14.35\%\pm 2.74$      \\
\emph{10 objects} &
	$5.82\%\pm 2.37$  & $6.46\%\pm 1.79$ &  $5.00\%\pm 2.62$  & $5.92\%\pm 2.12$     \\
\emph{15 objects} &
	$3.72\%\pm 1.42$  & $4.00\%\pm 1.54$ &  $1.59\%\pm 1.31$  & $2.48\%\pm 1.05$     \\
	\bottomrule
\end{tabular}
\end{center}
\caption{Communicative success using messages `inferred' by assuming a systemic relation within $\argmin_i$/$\argmax_i$ message pairs. The `compositionality by addition' method assumes that \MESS{$c, \argmax_i$} = \MESS{$c, \argmax_j$} - \MESS{$c, \argmin_j$} + \MESS{$c, \argmin_i$}. The `compositional network' is an MLP trained to predict \MESS{$c, \argmax_i$} from the other three messages. Table values are object recovery accuracies averaged for all $i$.}\label{tab:composition_accuracy}
\end{table*}

Second, we note as others that the composition-as-addition assumption is disputable, both in general and in the original application case \citep{linzen-2016-issues, Chen2017}. To abstract away from this issue, we train a `composition network' (an MLP with 2 hidden layers of 64 ReLU units) on the task of predicting 
\MESS{$c, \argmax_i$} from \MESS{$c, \argmax_j$}, \MESS{$c, \argmin_j$} and \MESS{$c, \argmin_i$}, therefore letting it discover any function for mixing values, and not involving addition \emph{a priori}. We leave out one dimension $i_0$ from training, and feed Receiver with the message predicted by the `composition network' from \MESS{$c, \argmax_j$}, \MESS{$c, \argmin_j$} and \MESS{$c, \argmin_{i_0}$}. If the language was compositional, this predicted message should behave like \MESS{$c, \argmax_{i_0}$}, but we found that, as in the case of addition, the average communication accuracy for all taken-out parameters dropped dramatically (again, at least 24 percentage points drop).
Full results are in the right column of Table~\ref{tab:composition_accuracy}.

\subsection{Categorical perception}

Above we essentially propose an analysis of discreteness both in production and perception. This can lead to more psycholinguistic-like queries about these emergent languages. For instance, one may ask whether classical `Categorical Perception' (CP) effects obtain, whereby two messages at a short distance in the latent space may be discriminated easily if (and only if) they are on two sides of a categorical boundary for interpretation purposes (see \citealp{liberman1957discrimination}, and \citealp{DamperHarnad-CP} for early discussions in the context of neural architectures).

\begin{figure}[ht]
	\centering
	\includegraphics[width=\columnwidth]{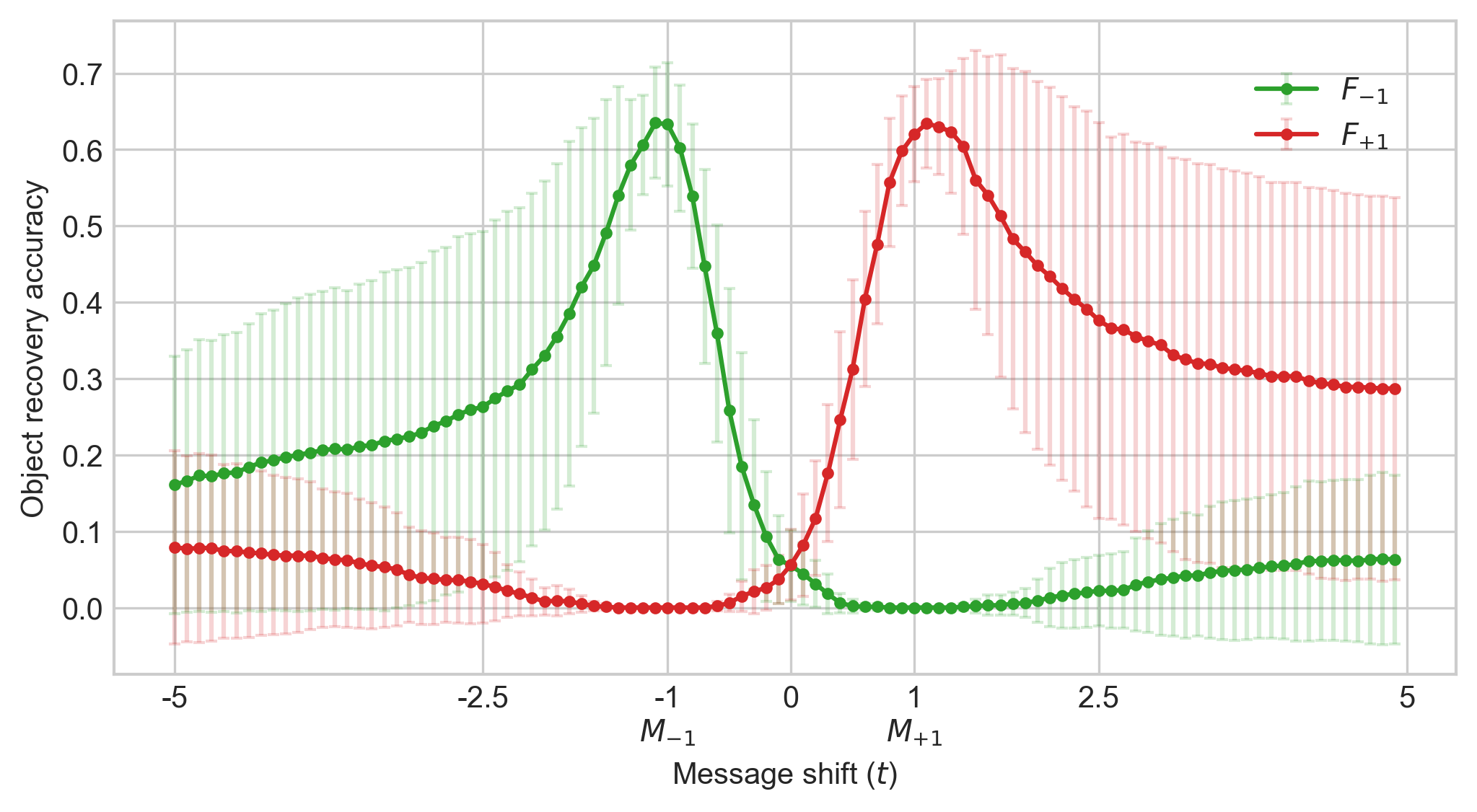}
	\caption{Categorical perception effect, demonstrated by accuracy of object recovery using messages shifted between two `meanings'.}
	\label{fig:categorical_perception}
\end{figure}

As an initial foray, we can investigate the sharpness of the boundaries of our discrete messages (i.e. distribution in latent space). For representation purposes, we sample pairs of messages, call them $M_{-1}$ and $M_{+1}$ generated by Sender for two choice functions $F_{-1}$ and $F_{+1}$. We explore a continuous spectrum of messages in the dimension connecting these two messages ($M_t = \frac{(1-t)M_{-1} + (1+t)M_{+1}}{2}$, continuously shifting from $M_{-1}$ to $M_{+1}$ as the continuous variable $t$ moves from $-1$ to $+1$). The messages $M_t$ are fed to Receiver together with contexts $C'$, and for each function $F_{-1}$ and $F_{+1}$ in turn, we calculate object recovery accuracy.  This is plotted in Figure~\ref{fig:categorical_perception} for an Extremity Game model trained in a strict, non-shared context setting with object size 5. The model shows that clusters have relatively sharp boundaries, especially in the direction of a message belonging to another cluster (the area where $x$ is between $-1$ and $+1$ in Fig.~\ref{fig:categorical_perception}).

We can thus identify a boundary around a cluster, and its width, providing the necessary setup to investigate CP effects: whether pairs of messages crossing such a boundary behave differently (e.g., are easier to discriminate) than a pair of equally distant messages both on one side of this boundary.

\section{Conclusion}

We propose a general signaling game framework in which fewer \emph{a priori} assumptions are imposed on the conversational situations. We use both production and perception analyses, and find that under appropriate conditions, which are met by most studies involving neural signaling games, messages become discrete without the analyst having to force this property into the language (and having to deal with non-differentiability issues). We find no evidence of compositional structure using vector analogies and a generalization thereof but do find sharp boundaries between the discrete message clusters.  Future work will explore other measures and alternative game settings for the emergence of compositionality, as well as more subtle psychological effects (Categeorical Perception) of continuous biological systems exhibiting discrete structure, like the auditory system.

\section*{Acknowledgments}

We acknowledge the funding support from ANR-17-EURE-0017, and greatly thank Marco Baroni, Diane Bouchacourt, Rahma Chaabouni, Emmanuel Dupoux, Roni Katzir, Philippe Schlenker, Benjamin Spector, Jakub Szymanik, and three ACL reviewers.


\bibliographystyle{acl_natbib}
\bibliography{acl2020}

\end{document}